\documentclass[conference]{IEEEtran}

\usepackage{amsmath}
\usepackage{booktabs}
\usepackage{amssymb}
\usepackage{graphicx}
\usepackage[caption=false]{subfig}
\usepackage{url}
\usepackage{array}
\usepackage{multirow}
\usepackage{rotating}
\usepackage{enumitem}
\usepackage{makecell}

\usepackage{amsthm}
\usepackage[linesnumbered,ruled]{algorithm2e}

\begin{document}

\title{Multi-Faceted Hierarchical Multi-Task Learning for a Large Number of Tasks with Multi-dimensional Relations}

\author{\IEEEauthorblockN{Junning Liu\IEEEauthorrefmark{1},
 Zijie Xia\IEEEauthorrefmark{1},
 Yu Lei\IEEEauthorrefmark{1},
 Xinjian Li\IEEEauthorrefmark{1} and
 Xu Wang\IEEEauthorrefmark{1}}
 \IEEEauthorblockA{\IEEEauthorrefmark{1} Tencent PCG\\
 Shenzhen,
 China \\ Email: \{korchinliu, zijiexia, mikeylei, xinjianli, caelanwang\}@tencent.com}
 }





\maketitle

\begin{abstract}
There has been many studies on improving the efficiency of shared learning in Multi-Task Learning(MTL). Previous work focused on the ``micro" sharing perspective for a small number of tasks, while in Recommender Systems(RS) and other AI applications, there are often demands to model a large number of tasks with multi-dimensional task relations. For example, when using MTL to model various user behaviors in RS, if we differentiate new users and new items from old ones, there will be a cartesian product style increase of tasks with multi-dimensional relations. This work studies the ``macro'' perspective of shared learning network design and proposes a Multi-Faceted Hierarchical MTL model(MFH). MFH exploits the multi-dimension task relations with a nested hierarchical tree structure which maximizes the shared learning. We evaluate MFH and SOTA models in a large industry video platform of 10 billion samples and results show that MFH outperforms SOTA MTL models significantly in both offline and online evaluations across all user groups, especially remarkable for new users with an online increase of 9.1\% in app time per user and 1.85\% in next-day retention rate. MFH now has been deployed in a large scale online video recommender system. MFH is especially beneficial to the cold-start problems in RS where new users and new items often suffer from a ``local overfitting" phenomenon. However, the idea is actually generic and widely applicable to other MTL scenarios.
\end{abstract}


\section{Introduction}\label{sec_intro}
Recommender systems (RS) are widely applied in online applications. To better model user preference and increase satisfaction, there has been an increasing trend of using Multi-Task Learning (MTL) to simultaneously predict various user feedbacks. With the forthcoming era of immersive short videos such as TikTok, Reels, Triller etc., this trend has been accelerated as in full screen mode there are more implicit parallel user feedbacks of more balanced importance v.s. in the list page era there is a dominant impression-click behavior thread. The mainstream RS ranking scheme is to apply MTL to predict various specific user behaviors and predefined task labels as accurate as possible, followed by a fusion model that aims to characterize the overall user satisfaction based on the specific output values of the MTL model, e.g. likelihood of thumbup, favoriting, sharing, completion ratio etc. 

Compared to Single Task Learning(STL), MTL is a general learning framework that learns multiple task simultaneously in one single model v.s. one model per task, it has not only been used in RS, but also widely applied in Natural Language Processing(NLP), Computer Vision(CV) etc. learning problems of various domains. MTL has the benefit of transfer learning and improves model generalization through induction bias. MTL also faces challenges such as negative transfer and seesaw phenomena\cite{tang2020progressive} induced by the complex correlation between tasks. Essentially, deep MTL networks needs to jointly accomplish representation learning and information routing between tasks\cite{tang2020progressive}. There has been many research that focus on improving the shared learning efficiency between 2 tasks or small task groups. From the simplest hard sharing to MMOE\cite{ma2018modeling} and PLE\cite{tang2020progressive}, with innovative shared learning structures, these works try to improve the shared learning efficiency and better address the confliction between tasks. The focus is on the micro scale task cooperation for small task groups.

On the other hand, there are often much more tasks beneficial to model together in real world applications. For example in immersive short-video applications, users generate richer and more subtle feedbacks such as time spent, likes, subscription, comments, sharing etc., and many predefined labels such as 3s skip (swipe away within 3 seconds), 80\% completion ratio etc. Joint learning these tasks with MTL has the benefit of better representation learning and more shared learning among tasks. Furthermore, in RS and many other applications, it is frequent to have different value ranges of certain properties exhibiting very different correlation patterns with the predicted label. Normally these property values are simply treated as sample features to the model, however, this often results in overfitting in certain feature regions especially when we have imbalanced and insufficient data samples. Despite that these feature regions may lack data samples, it is often important to improve the model's accuracy there as they can be regions of higher business value, or regions important for the long term value of growth, content ecosystem etc., e.g. new users, new content facing the cold start situation. Another approach is to split the different user/item/context regions into independent tasks, then there will be a cartesian product style increase in the number of tasks as these are orthogonal dimensions that independently divide the samples. 

The increasing scale of task numbers in MTL poses new research challenges. Current MTL technologies do not scale well with the task number increase and still lack sufficient exploration in this area. This explains why we observe the common industrial practice of employing 2 or even 3 MTL models in the ranking service to handle total of 10-20 tasks where each MTL model normally handles 2-6 tasks. However, this practical approach is essentially a compromise between MTL and STL.  

Addressing large number of tasks in MTL is challenging. These tasks could have complex relations and are highly imbalanced as a large fraction of these tasks could have little data. The MTL learning structures in previous works such as MMOE, PLE handle well with two or small number of tasks. With a large number of tasks, we will show later that simply plugging in all tasks with flat sharing using these structures can hardly yield significant gains compared to not adding these tasks. We need new MTL models that exploit the semantic correlation structures among the tasks. Thus, designing an efficient MTL structure that can support efficient shared learning with large number of correlated tasks has becoming a critical aspect for MTL recommender system. Meanwhile, advancing MTL's scalability also has great value for the generality of MTL as a universal learning framework.

In this work, we propose a noval Multi-Faceted Hierarchical multi-task learning model(MFH) that aims at providing a more efficient multi-task learning for a large number of tasks through scalable efficient cooperative learning designs. The main contribution of the paper is summarized as follows: 
\begin{itemize}
\item \textit{Micro} and \textit{macro} perspectives of network design for cooperative learning in MTL are brought to attention for better understanding of the previous efforts on MTL model design and what is demanded for better scalability. Also a \textit{local overfitting} phenomenon is described and clarified. These concepts are important notions to understand the core issues of scalable MTL.    
\item A new MTL model MFH is proposed to better address the challenge of scalable efficient multi-task learning with three major characteristics: Multi-Faceted, Hierarchical and Heterogeneous. The multi-faceted and hierarchical design combined together introduces multidimensional implicit induction biases and results in a much more efficient and sufficient representation learning, thus greatly relieves the local overfitting and the data scarcity issue with smaller tasks. In addition, MFH network is more heterogeneity-friendly and provides great flexibility for the model to better customize the tasks and generate further improvement. 
\item Extensive experiments are carried out to evaluate the MFH in both a large-scale industrial short video app and a public benchmark dataset. Offline evaluation shows that MFH outperforms the baseline of a flat 9-task model by 0.46\% in MSE. Online A/B test results in WeSee APP also demonstrate the significant improvement of MFH over SOTA MTL models in real-world applications, with 2.14\% increase in app-time per user, 0.19\% increase in retention rate. The improvement is much more significant on new users as 9.10\% in app-time per user and 1.85\% in retention rate. MFH has been successfully deployed in our recommender system now and generates significant business values, especially for the cold-start recommendation. 
\item Furthermore, paired up with a generic fusion algorithm, MFH has a greater value of serving as a component of a universal ranking that unify the ranking for heterogeneous candidates in a broader context, e.g. mixed ranking with organic video content,ads and live broadcast etc. 
\end{itemize}

\section{Related Work}\label{sec_related_work}

\subsection{Multi-Task Learning Models}
Multi-Task Learning\cite{caruana1997multitask,ZhangYangMTLSurvey} is a general learning framework that improves the model generalization through cooperative learning between tasks. It explores the commonalities and differences between different tasks to facilitate the joint learning. MTL has been successfully applied to a wide range of applications, from RS\cite{Davidson10theyoutube,tang2020progressive,bansal2016ask,felfernig2018group} to NLP\cite{collobert2008unified,sanh2019hierarchical} and CV\cite{nguyen2019multi,fan2017hd,kendall2018multi} etc.

\subsection{Shared Learning in MTL}
There are many studies on improving the shared learning efficiency in MTL. Shared-bottom\cite{caruana1997multitask} is the first simple structure for task sharing which has its limit on task conflict. Cross-Stitch Network\cite{misra2016cross} and Sluice Network\cite{ruder2017sluice} both learn static weights of linear combinations to fuse representations for different tasks selectively. Then several approaches apply the gate structure and attention mechanisms to model the sample dependent correlation between tasks. MOE\cite{jacobs1991adaptive} first introduces expert modules and uses gating network to fuse the expert outputs for upper task towers. MMOE\cite{ma2018modeling} extends MOE to utilize task-specific gates to provide customized fusion for each task. M3oE\cite{xie2021real} extends the customized gates in MMOE with multi-head gates. PLE\cite{tang2020progressive} further improves the shared learning efficiency by differentiating task-specific experts and shared experts, and adopting a progressive routing mechanism. On top of PLE, MSSM \cite{ding2021mssm} trains a field-level sparse connection to provide more flexible feature combination for different tasks, and replace the gates with a matrix multiplied by a mask vector.

All of the works mentioned above focus on tasks of small groups, and are often evaluated with 2 tasks as an example. There still lacks research studying the structures of the task relation graph for a large number of tasks. Recently, ESMM\cite{ma2018entire} constructs joint loss of CTCVR based on CTR and CVR's task relation. \cite{wen2021hierarchically} generalizes ESMM to model the task relations with a Bayesian graph and construct joint losses as in ESMM along the paths in the graph, it also introduces more tasks. These works exploit task relation structures in loss design, instead of the network structure design for shared learning. To the best of our knowledge, this work is the first to address this for MTL in RS.

Besides, another thread of research has been applying neural architecture search(NAS)\cite{zoph2016neural} and other AutoML methods to learn efficient MTL architectures automatically. Prior works such as \cite{rosenbaum2017routing}, SNR\cite{ma2019snr},MTNAS\cite{chen2020boosting} etc. are all efforts in this direction. However, this is still in early stage as the search range of network structures is often based on certain simplified assumptions and the learning cost is expensive.

\subsection{Local Overfitting}
Overfitting is one of the most common problems in machine learning. Usually, it is manifested as a large performance gap between the training and testing set, with a much worse performance on the testing set. However, as mentioned in \cite{lawrence2000overfitting}, overfitting can vary significantly throughout the input space. \cite{jiang2006svm} first mentioned a term of partial overfitting and pointed out certain data distributions are prone to lead to partial overfitting.  \cite{webster2019detecting} also mentioned local overfitting on different image patches. Despite these mentions, local overfitting has not been officially introduced to the best of our knowledge. The traditional methods to address overfitting normally do not apply well with local overfitting as it is often caused by drastic feature-label pattern change combined with highly imbalanced training samples across the feature regions. Thus how to alleviate local overfitting effectively is a challenging issue worthy of study. In this work, we formally introduce local overfitting and propose MFH to address it.

\section{Preliminary} 
\subsection{MTL Ranking in an industrial Recommender System}
First of all, we briefly introduce a real-world recommender system in WeSee, a short-video playing APP of Tencent, which serves tens of millions of users every day for immersive video watching experience. 
For a user request, the recommender system works to generate a recommended list of videos from a ten-million-scale candidate pool, then present the recommendations on the user's mobile screen one video at a time.  
Each recommended video will start autoplay and the user may take various actions such as keep watching, swipe, like, comment, share, etc.
In particular, if the user swipe up the current video, the system will show the next video from the recommended list. 
Upon all videos of the recommended list are presented, a new request will be triggered to generate another list. 
The goal of the system is to recommend favorable videos that maximize the user satisfaction which is normally quantified by total app time.

As in most industrial recommender systems \cite{covington2016deep}, our recommender system adopts a two-stage design that contains two core processes, candidate recall and ranking. 
In this paper, we focus on the ranking part, which aims to rank thousands of recalled candidate videos and select the top ones for presentation. 
As the widely adopted RS ranking framework shown in \cite{tang2020progressive}, our ranking system is composed of two parts: an deep MTL model and a Evolution Strategy(ES)/ Reinforcement Learning(RS) fusion model. The MTL model jointly outputs the likelihood of various user behaviors and its goal is to predict those concrete signals as accurate as possible. Based on user and context state features, the ES/RL fusion model outputs hyper parameters for a proxy function which calculates the final fusion score to rank the candidate videos, the proxy function takes the MTL model's outputs as its input. The goal of the fusion model is to synthetically characterize the user satisfaction score based on the concrete signal that the MTL model predicts. In our practice, there are two MTL models each deals with a different group of tasks. We will first use the play task group as an example.

\subsubsection{Play Task Group MTL}
In the play task group MTL, we focus on three important tasks that are highly related to the video watch time, i.e., \textit{Play Completion Ratio} prediction, \textit{Play Finish Rate} prediction, and \textit{Play Skip Rate} prediction. 
For simplicity, we will denote the three tasks by \textit{Cmpl}, \textit{Finish} and \textit{Skip} respectively in the rest of this paper. 

Specifically, Cmpl task is a regression task that predicts the completion ratio of a video view. 
The label is defined as:
\begin{align}
y_{cmpl}=\frac{watch~time}{video~length}
\end{align} 
The watch time of a video view may often exceed the video length due to re-watch and the auto-replay nature of immersive short-video play APPs, thus $y_{cmpl}\in[0,\infty)$. 
To handle exception cases, we truncate $y_{cmpl}$ to ensure its maximum value is below a certain threshold.

Finish task is a binary classification task that predicts the probability of watching a video to the end. 
\begin{align}
y_{finish}=
\begin{cases}
1, &\text{if}~watch~time \geq video~length\\
0, &\text{otherwise}
\end{cases}
\end{align} 
Skip task is a binary classification task that predicts the probability of quick skipping a video within a short time.
\begin{align}
y_{skip}=
\begin{cases}
1, &\text{if}~watch~time \leq c~\text{seconds}\\
0, &\text{otherwise}
\end{cases}
\end{align} 
where $c$ is a small constant number. 

The three tasks model users' watching behaviors from slightly different perspectives. 
The Cmpl task reflects a user's commitment to a video continuously, while the Finish and Skip tasks focus on modeling users' positive and negative viewing experience, respectively. In our system, the three tasks are jointly learned with one MTL model.  

\subsection{Local Overfitting in Cold Start}
In our practice a \textit{local overfitting} phenomenon is observed. By local overfitting, we mean that the model is overfitted in part of the input space or the degree of overfitting in parts of the input space is much more serious than other regions. For example, there are three different groups of users in our App: \textit{new users}, \textit{low-activity users} and \textit{high-activity users}, activity level is determined according to total video watch time of the user. Naturally there are far less data samples for new/low-activity users than for high-activity users. In our MTL ranking, all of the baseline models exhibit overfitting on all tasks for new/low-activity users, in contrast to high-activity users. Intuitively, this is because the training samples of new/low-activity users are much fewer than that of high-activity users and the model parameters trained with such imbalanced data is dominated by the data pattern of high-activity users.
Thus the predictions on new/low-activity users are negatively affected by high-activity users, leading to unsatisfactory recommendation results and user experience. However, cold-start - addressing the recommendation for new users or new items, is a problem of great business value for retention and growth. Models are often developed separately for cold-start in industrial practice. This is far from an ideal approach as the new user model still suffers from few data samples, additional training/serving cost, and more importantly the common patterns shared between different user groups can not be transferred to benefit the minority groups. 

Furthermore, this local overfitting phenomenon is not only restricted to cold start problem of new users. It is actually quite common to encounter. New items, new users, minority context, ... , in general any feature value regions that exhibit different pattern from the main regions and have much less data samples may suffer from local overfitting.

A natural idea to address local overfitting is to split original task's feature regions into independent tasks with the overfitting regions as separated tasks. For example, we can split the Cmpl task into three new tasks: New\&Cmpl, Low\&Cmpl, High\&Cmpl, representing Cmpl for new users, Cmpl for low-activity users and Cmpl for high-activity users respectively. This will give the overfitting regions more customized care and hopefully relief the overfitting. However, this also introduces new challenges as the number of tasks increases from 3 tasks to 9 tasks in a cartesian product fashion. With a large number of unbalanced tasks that have multi-aspects of correlations, is the traditional MTL structure well prepared to scale efficiently? This is a natural question to ask. We will demonstrate later in  Table 1 of the Experiments Section that addressing many more tasks with flat sharing can hardly bring any substantial gains. Next, before showing our solution with the design of the Multi-Faceted hierarchical MTL model in the next Section, we will first explain the perspectives of micro and macro shared learning structures in MTL and introduce the baseline models from these perspectives. 


\subsection{Micro and Macro perspective of cooperative learning in MTL} 
\subsubsection{Micro level cooperative learning structure - MTL Switcher}
\begin{figure}[t]
\centering
\includegraphics[width=0.99\columnwidth]{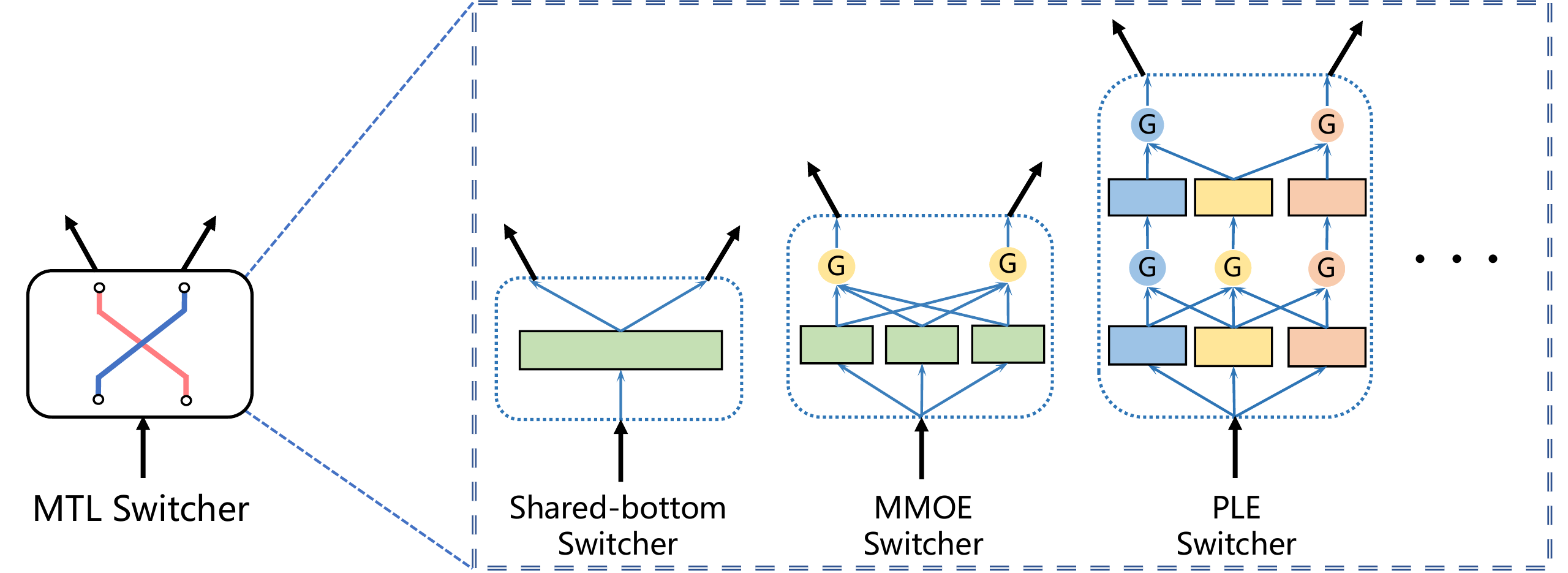}
\caption{The MTL switcher and its possible instantiations.}
\label{fig_switcher}
\end{figure}

To illustrate the micro level coordinated learning structures in MTL, we introduce a concept called \textit{switcher} in deep learning models. As depicted in Fig. \ref{fig_switcher}, a switcher is a neural architecture that takes one input and branching out multiple ($\geq 2$) latent outputs. The input of the switcher can be any types of features, embeddings or intermediate latent representations in the network, and its outputs are some latent representations that will be fed into upper-level networks such as any hidden layers in the network or particularly specific task towers in deep MTL models. As discussed in \cite{tang2020progressive}, MTL needs to jointly address representation learning and information routing. Abstractly, switchers can be used to deal with the micro level cooperative learning of diverting one input to multiple intermediate latent outputs. As we see in Fig. \ref{fig_switcher}, MTL structures such as shared-bottom, MMOE, PLE are just switchers of different types. \footnote{Note that in the shared-bottom case the bottom layer could downgrade to zero layer as the hard sharing scheme which simply shares the input directly to the branches.}  Previous MTL research has been focusing on switcher innovations improving the micro level cooperative learning efficiency, partially because we were not dealing MTLs with a large number of tasks.


\subsubsection{Macro level cooperative learning structure}
\begin{figure*}[t]
	\centering
	\hspace*{\fill}
	\subfloat[baseline 3-task model\label{fig_baseline_3task}]{
		\includegraphics[height=0.15\textheight]{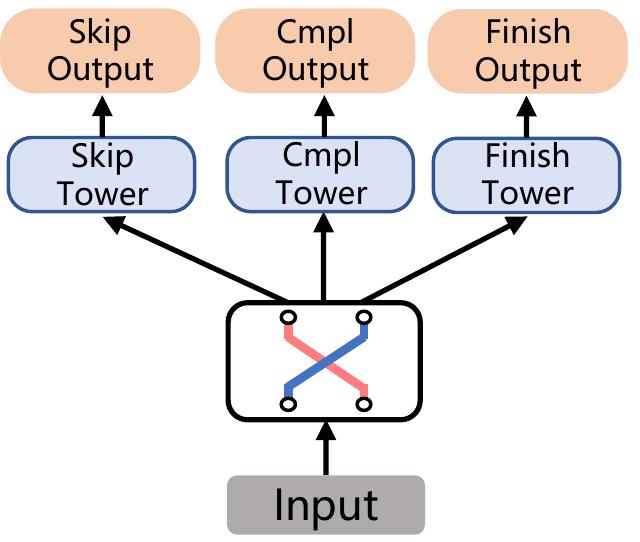}}
	\hspace*{2em}
	\subfloat[baseline 5-task model\label{fig_baseline_5task}]{
		\includegraphics[height=0.15\textheight]{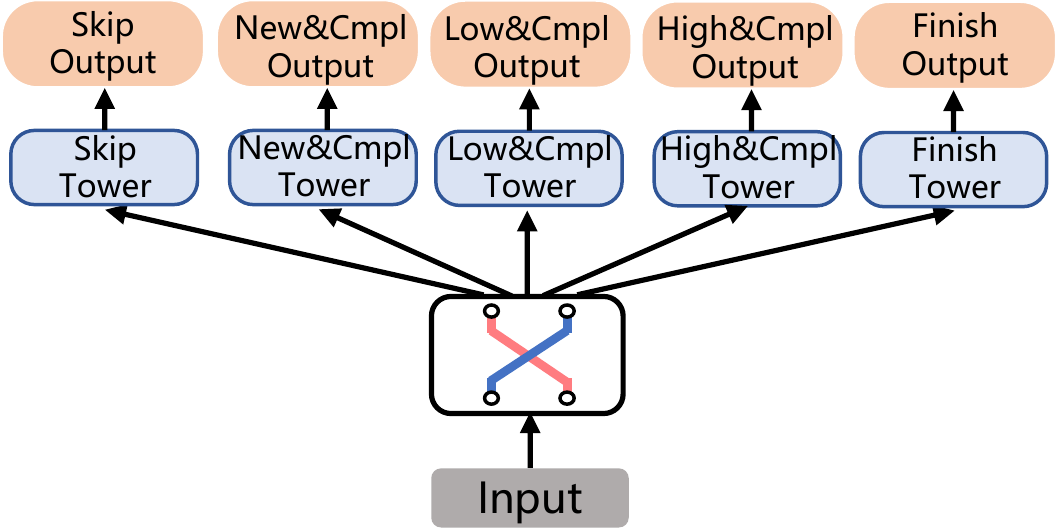}}
	\hspace*{\fill}
	\newline
	\hspace*{\fill}
	\subfloat[baseline 9-task model\label{fig_baseline_9task}]{
		\includegraphics[height=0.16\textheight]{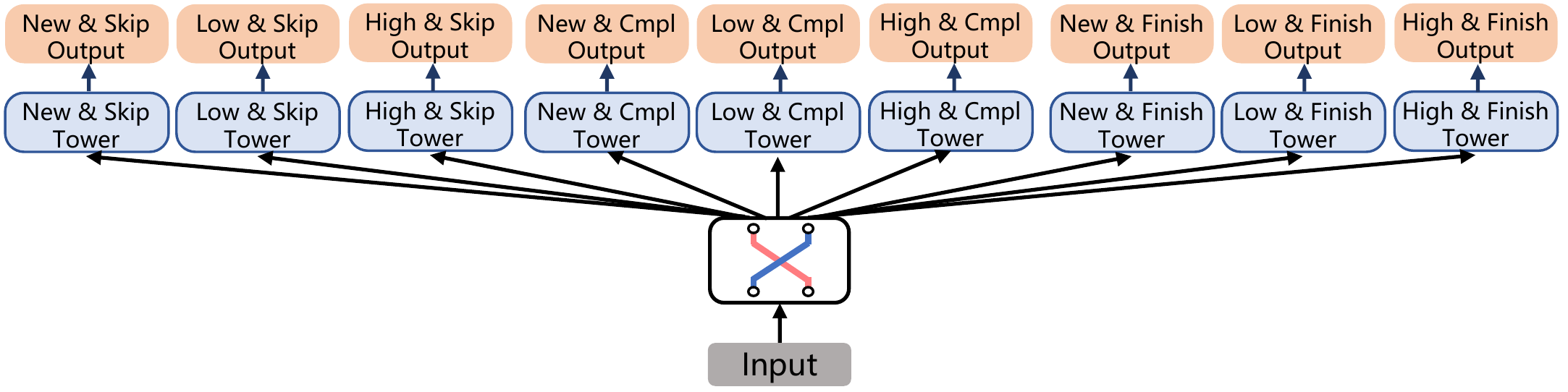}}
	\hspace*{\fill}
	\caption{Baseline MTL models.}
	\label{fig_result_T}
\end{figure*}

As discussed earlier the number of tasks can easily increases to a large number. This raises the importance of the macro level cooperative learning structure for MTL models. The macro learning structure concerns with the macro scale information sharing among tasks. One straightforward macro learning structure will be a flat branching structure as shown in fig. \ref{fig_result_T}, an opposite alternative can be a chain structure like the asymmetric sharing described in \cite{tang2020progressive}, other possibilities can also be a tree etc. For the same macro level structure, any micro level switchers can be used for the local diverting, we will examinate different combinations of micro switchers for each macro structure candidate and use the performance of the best switcher choice to represent the macro structure. And in general we find that it is normally better to use the SOTA switchers everywhere for a macro structure given sufficient training data.

\subsection{Baseline Models} 
We choose the straightforward macro structure of flat sharing for the baseline models and there are a few extended variations.

\subsubsection{Baseline 3-task Model}
Figure \ref{fig_baseline_3task} illustrates a baseline flat 3-task model with one switcher and three towers that corresponds to the three tasks Skip, Cmpl, and Finish, respectively. 
Note that the baseline 3-task model may have different versions, depending on the specific switcher architecture it adopts\footnote{In the rest of this paper, we will describe other MTL models using abstract switchers in the same way.}. In our case when the switcher upgrades from simple to more complex SOTA ones, i.e., Shared-bottom $\rightarrow$ MMOE $\rightarrow$ PLE, corresponding performance improvements are observed.

However, in spite of different switchers, the baseline 3-task model cannot well address the new user cold-start problem mentioned earlier which shows local overfitting is beyond the scope of micro level switchers. 

\subsubsection{Flat 5-task and 9-task Models}
To alleviate the new user issue, we attempt to further divide a prediction task into three sub tasks according to the division of user groups, i.e., prediction on new users, prediction on low-activity users, and prediction on high-activity users. 
The idea is to improve the predictions on new/low-activity users by allowing them more independent optimization without loss of accuracy on high-activity users. To validate this idea, we implement a partly-divided baseline flat 5-task model (Fig. \ref{fig_baseline_5task}) by dividing only the Cmpl task into three sub tasks New\&Cmpl, Low\&Cmpl and High\&Cmpl, and a fully-divided baseline flat 9-task model (Fig. \ref{fig_baseline_9task}) by dividing all tasks in the same way. 
Here, the notation New\&Cmpl denotes the prediction task of Cmpl on New users, and other notations have similar meanings accordingly. 
Later, offline evaluation results of Table 1 in the Experiments Section show that the flat 5-task and 9-task models make limited improvement compared to baseline 3-task model in terms of prediction accuracy on new/low-activity users, and the local overfitting is still significant.
Next, we illustrate the MFH model which is essentially designed with a nested multi-hierarchical tree structure for solving this problem.  

\section{Multi-Faceted Hierarchical Multi-Task Learning}\label{sec_method}
In this section we first introduce the concept of \textit{facet} for tasks, then for sake of readability, we employ the aforementioned 9-task problem as an example to describe our proposed models. Last, we generalize our models to a general $N$-faceted multi-task problem setting.

\subsection{Facets of Tasks}

\begin{figure*}[t]
\centering
\includegraphics[width=0.85\textwidth]{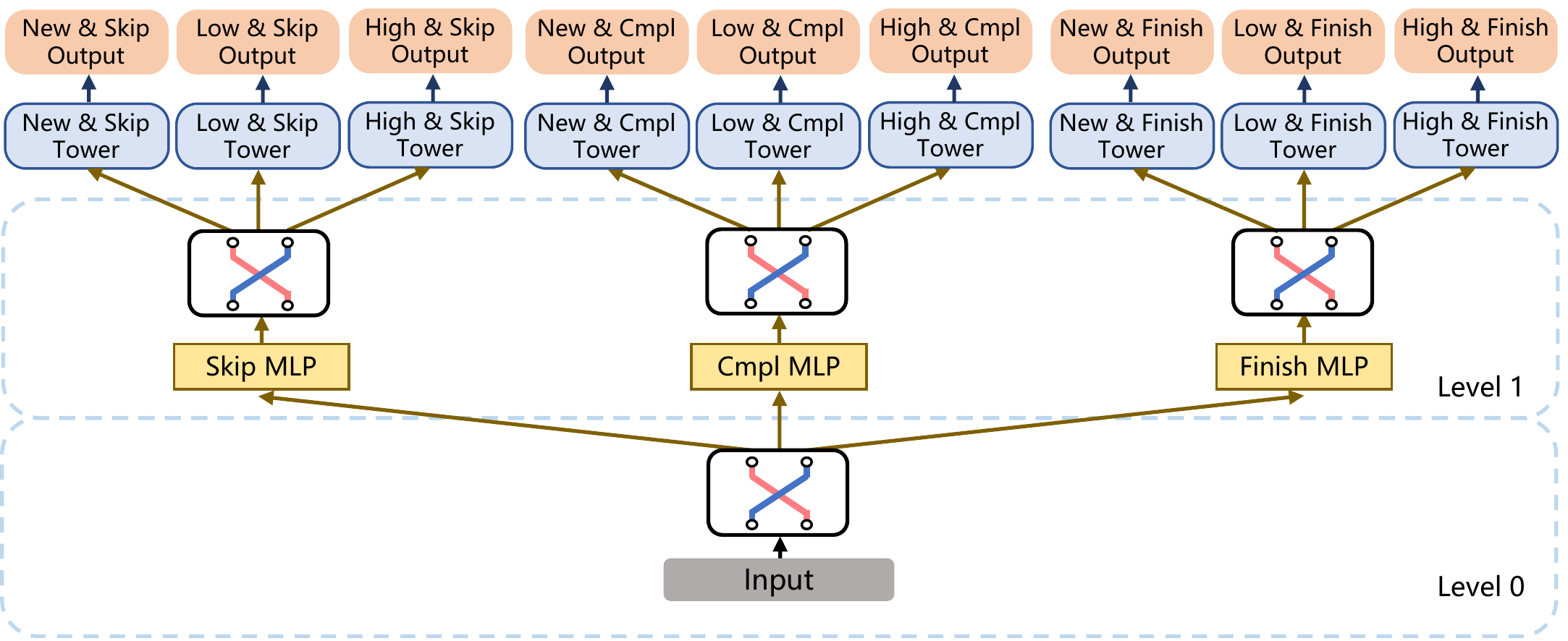}
\caption{Hierarchical MTL (H-MTL) Model.}
\label{fig_hmtl}
\end{figure*}

We introduce a concept of \textit{facet} for tasks, facets are orthogonal dimensions that every task has. There are several partitions for each facet that can divide tasks into groups. For example in the 9-task problem setup, each task simultaneously has two \textit{facets}, i.e., \textit{user behavior facet} = \{\text{Cmpl,~Finish,~Skip}\} and \textit{user group facet} = \{\text{New,~Low-activity,~High-activity}\}, where each facet contains three \textit{partitions}. 
The combination of any two partitions from the two facets defines a specific task. Facets provide prior structures for the correlation between tasks, there are correlations along each facet's aspect and tasks share common facet partitions have stronger correlations. In the 9-task MTL, each task correlates more with two tasks sharing the same partition on one facet and other two tasks on another facet. 
For example, Task New\&Cmpl shares the common user group of new users with New\&Finish and New\&Skip, and shares the common user behaviour of complete watching with Low\&Cmpl and High\&Cmpl, at the same time. In general, the number of facets could be three or even more, e.g., the popularity of the video or the video length can each be another new facet. The number of tasks increases very quickly in a Cartesian Product fashion as more facets are introduced.

This type of multi-facet multi-task problem is actually very common in industrial practice. In spite of its universality, few research has studied the problem of macro level cooperative learning strategies, partially because the community has been focused on innovating micro level structures such as MMOE, CGC and PLE. With the switcher modules sharpened, it is a good time to address the challenges on how to scale the cooperative learning in the multi-facet MTL setup from the macro perspective.

\subsection{Hierarchical MTL}
In this subsection, we introduce a Hierarchical MTL (H-MTL) model for the 9-task problem. 
As depicted in Fig. \ref{fig_hmtl}, H-MTL utilizes a two-level tree architecture to model the task relationships in both facets and share the facet latent representations between tasks in a hierarchical fashion.
At level 0, a switcher is adopted to learn the task relationship in the user behavior facet based on the input features, which connects to three MLPs(Multilayer Perceptrons) at level 1 that correspond to three partitions of the user behavior facet: Skip, Cmpl and Finish, respectively. 
Each MLP outputs a hidden representation to feed a MTL switcher which learns the task relationship in the user group facet conditioned on a particular partition of the user behavior facet, and connects to three task tower networks that corresponds to the combination of this user behavior and one of the three user groups (i.e., new, low-activity and high-activity). 
Each task tower network concentrates on the corresponding task, and predicts the final score for that task. 
Formally, the output of a specific task, e.g., Task New\&Skip, can be abstractly formulated as:
\begin{align}
&Output_{New\&Skip}=Tower_{New\&Skip}(Switcher_{Skip}^{New\&Skip}( \nonumber\\ 
&\quad MLP_{Skip}(Switcher_{Input}^{Skip}(Input))))
\end{align} 
where $Switcher_{X}^{Y}$ indicates the corresponding output of $Switcher_{X}$ for (hidden) Task $Y$, and in general the MLP can downgrade to zero layers in which case the lower level switcher will feed directly to the upper level switchers.

\begin{figure*}[t]
\centering
\includegraphics[width=0.85\textwidth]{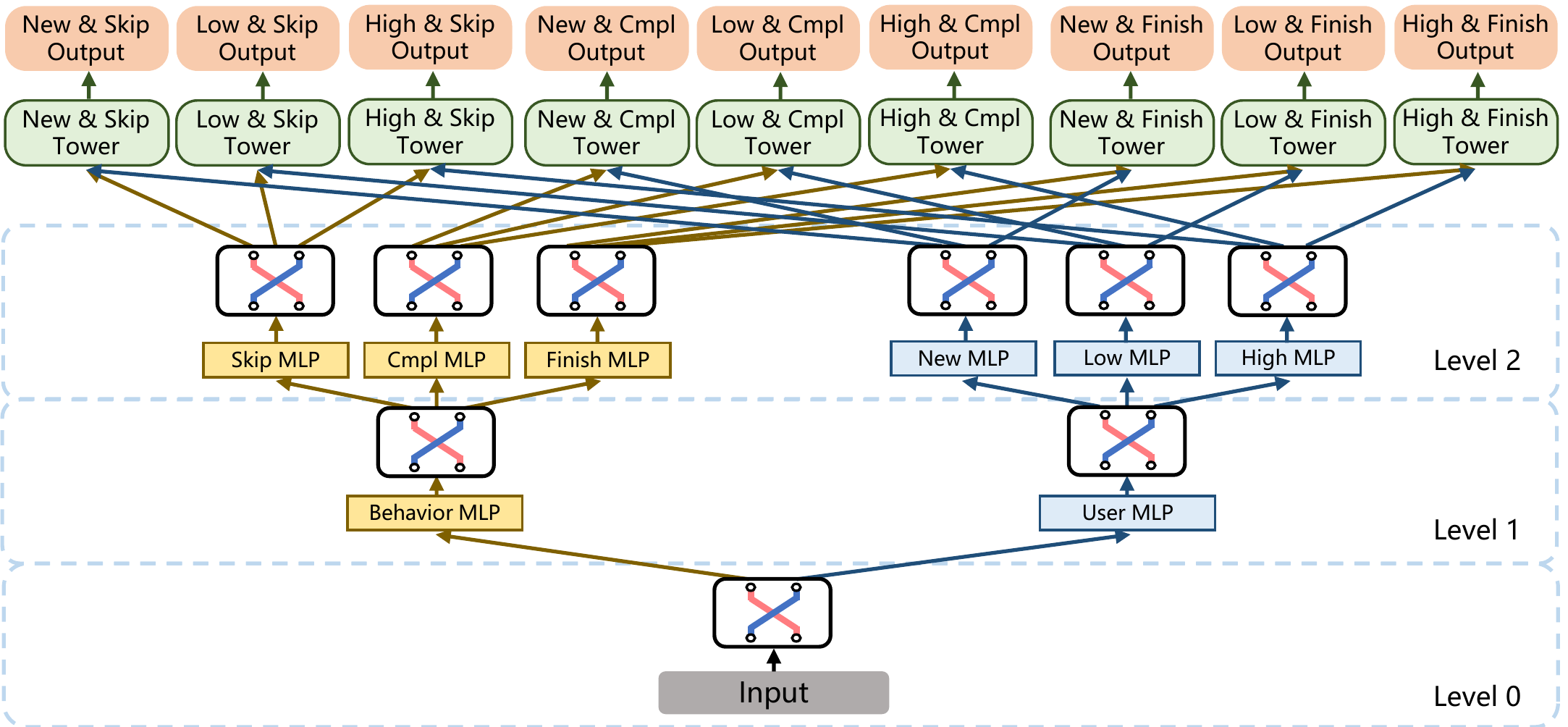}
\caption{Multi-Faceted Hierarchical MTL Model(MFH).}
\label{fig_mhmtl}
\end{figure*}

From the macro cooperative learning perspective, H-MTL avoids to branch out directly from the input to all tasks as the baseline flat 9-task model does. 
Instead, it adopts a hierarchical structure for a multi-level tree sharing among tasks. At each level, switchers are used to branch out semantic representations for the upper level sub-trees. In general, the tree starts from level 0 to level $k \leq N-1$ when we have $N$ facets. Different permutations of the facets form different trees. For example, we can divide by user groups first then further divide by user behaviours.

\begin{figure*}[t]
\centering
\includegraphics[width=0.85\textwidth]{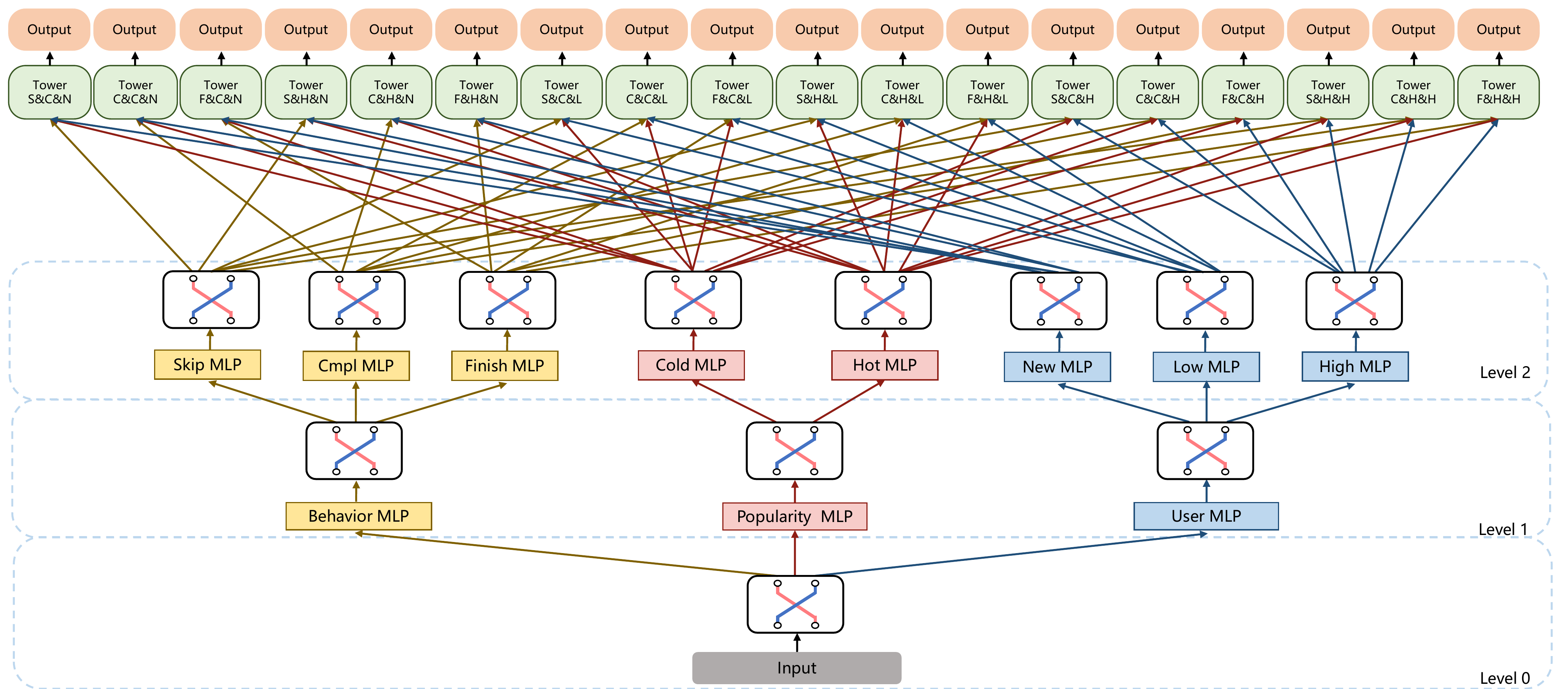}
\caption{3 Facets MFH Model.}
\label{fig_3facets}
\end{figure*}

\begin{figure*}[t]
\centering
\includegraphics[width=0.85\textwidth]{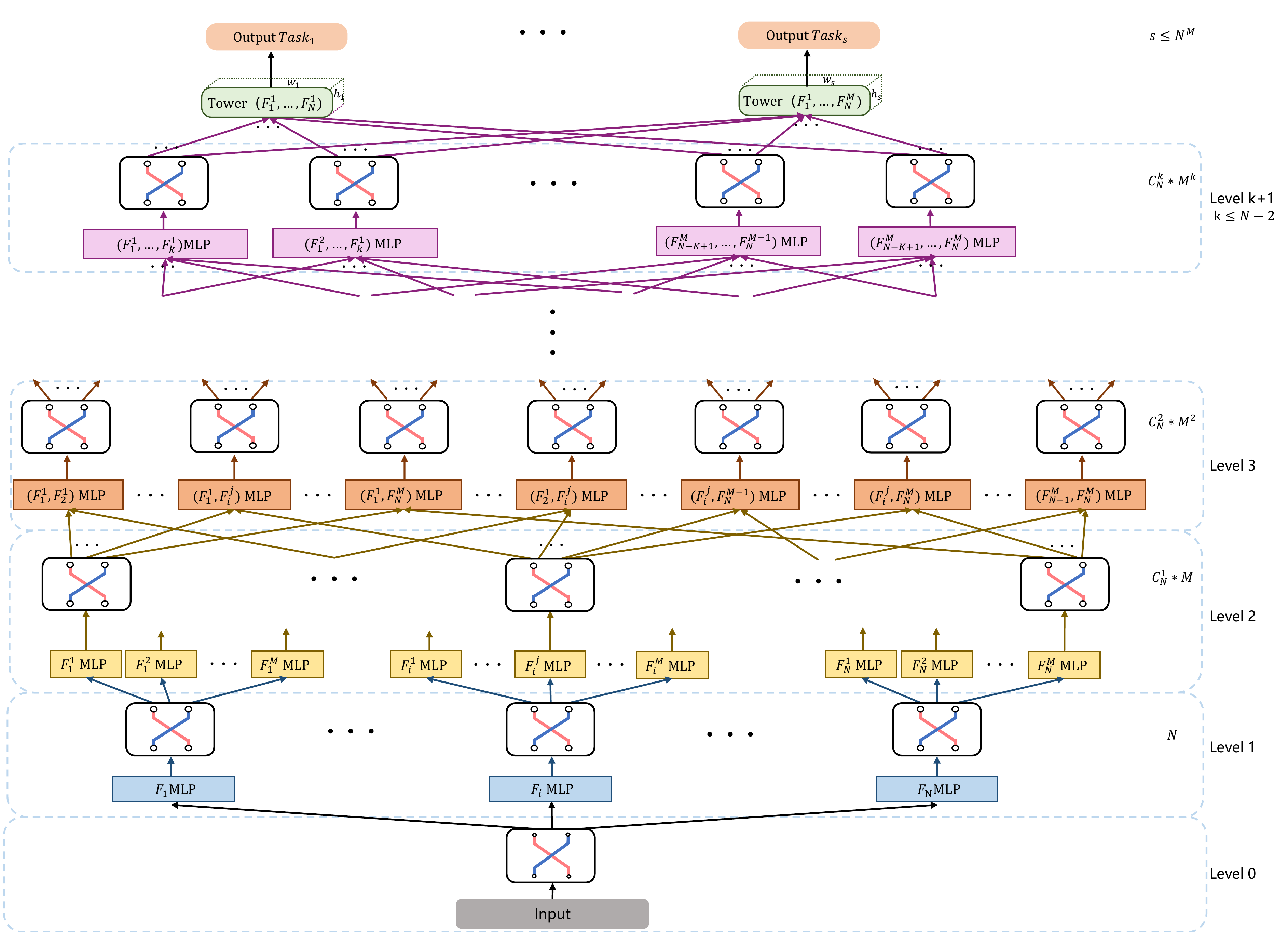}
\caption{Generalized Version of the MFH Model.}
\label{fig_mhmtl_general}
\end{figure*}

\subsection{Multi-Faceted Hierarchical MTL (MFH)}
Although the H-MTL model captures the hierarchical task relationships with multi-facets, we have to choose one particular tree corresponding to one specific permutation of the facets. However, the facets are normally important explicit dimensions along which the tasks lie strong correlation, they are orthogonal aspects of parallel importance. 

To further improve the efficiency of information sharing and cooperative learning among the tasks, we propose a more comprehensive model, named Multi-Faceted Hierarchical MTL (MFH). 
MFH is essentially composed of multiple H-MTL trees that are nested together. With the 9 task problem as an example, as shown in Fig. \ref{fig_mhmtl}, at level 0, the switcher network learns the inter-facet task relationship between two facets, and branches out to the two facets' MLPs at level 1. The upper-level structures of MFH can be simply regarded as the combination of two variants of the H-MTL model. 
In particular, each tower network linearly combines the hidden outputs from two different paths connected to the input, and outputs the predicted score for a specific task. 
For example, the output of Task New\&Skip can be abstractly formulated as:
\begin{align}
&Output_{New\&Skip}=Tower_{New\&Skip}(\nonumber\\ 
&\quad Switcher_{Skip}^{New\&Skip}(MLP_{Skip}(Switcher_{Behavior}^{Skip}(\nonumber\\ 
&\quad\quad MLP_{Behavior}(Switcher_{Input}^{Behavior}(Input)))))~+\nonumber\\ 
&\quad Switcher_{New}^{New\&Skip}(MLP_{New}(Switcher_{Group}^{New}(\nonumber\\ 
&\quad\quad MLP_{Group}(Switcher_{Input}^{Group}(Input))))))
\end{align} 
where $Switcher_{X}^{Y}$ indicates the corresponding output of $Switcher_{X}$ for (hidden) Task $Y$, and $+$ denotes the operation of linear combination. 

Compared to previous H-MTL model, MFH further improves the learning efficiency in the following aspects. 
First, it models three-fold task relationship by extending to three levels of MTL switchers, i.e., the inter-facet relationship, the first-order intra-facet relationship, and the second-order intra-facet relationship in the context of a particular partition of another facet. 
Second, it enables each task to share semantic information simultaneously with multiple sets of strong correlated tasks according to the shared facet partitions. With MFH, we can expand to any number of orthogonal trees of different facet permutations and nest the intermediate and leaf nodes as they cross. This maximizes the representation learning through multi-dimensional intersecting paths that support more shared learning.

\subsection{Generalization to $N$-faceted Multi-Task Setting}
In this subsection, we generalize the MFH model to a generic $N$-faceted multi-task problem setup.
Let $F_i$ denote the $i$-th facet, and $F_i^j$ denote the $j$-th partition of facet $F_i$, facet $F_i$ contains $M_i>1$ partitions. For simplicity of presentation we assume all facets have equal $M$ partitions, and it is easy to extend the following description to the general case.   
Then, there would be in total $s \leq M^N$ tasks in this multi-task setting.
Each task is associated with one element of the Cartesian product of the $N$ facets, denoted by a $N$-tuple, $(F_1^{j_1},..., F_N^{j_N})$, where $j_1,...,j_N \in \{1,...,M\}$.
Given these definitions, we illustrate the generalized version of MFH model in Fig. \ref{fig_mhmtl_general}. 

Specifically, the level 0 switcher (i.e., the root node) expands the N facets MLPs $F_i$ MLP $i$ from 1 to N. The level 1 switchers expand each facet to its M partition MLPs. In general, for any $k < N-1$, level $k$ contains $C_N^{k-1}*M^{k-1}$ switchers and they expand to $C_N^k*M^k$ MLPs and switchers at level $k+1$. Each MLP at level $k+1$ has a unique code formed by a combination of $k$ unique facets each with a specific partition. Upper MLPs and lower MLPs are connected through the lower level switchers if the upper level MLP's code contains the lower level MLP's code as a subcode. 
In general, we can choose to expand the multi-facet network to any level of $k$, $1<k<N-1$, then connecting directly to the $s \leq M^N$ towers for output tasks. We can have at most $M^N$ tasks but do not necessarily need to split all the tasks if there is not high business value or special pattern for the considered input regions. 

Fig. \ref{fig_3facets} shows a 3-facets MFH of 18 tasks for the Play Task Group with an additional facet of video popularity, differentiating new items and old ones. This jointly models the complete cold-start problem for both new users and new items. 

\subsection{Heterogeneity of MFH}
MFH is more heterogeneity-friendly than flat MTL. In Fig. \ref{fig_mhmtl_general}, MFH's task towers and MLPs can all be designed heterogeneously. The size of the task towers can be customized to better fit the special task. For example, we can use smaller size MLP layers for task towers with less training samples. And we can do this for the intermediate MLPs that corresponds to tasks of less data samples as well. MFH is more flexible on heterogeneity as the common shared root is thinner and there are various granularities of sharing that can be customized to be heterogeneous. In addition to the structure, different tasks can also have customized input features only available for themselves.

Paired up with a generic fusion algorithm, MFH's heterogeneity has a greater value of serving as a component of a universal ranking that unify the ranking for heterogeneous candidates in a broader context, e.g. mixed ranking with organic contents, ads and live broadcast etc.

\section{Experiments} \label{sec_experiments}
In this section, offline and online experiments are performed on both a large-scale industrial recommender system and a public benchmark dataset to evaluate the effectiveness of the proposed models.

\subsection{Evaluation on an industrial Video Recommender System}
In this subsection, the proposed models are evaluated with a large-scale online video recommender system.
\subsubsection{Dataset} We collect an industrial dataset through sampling user logs from Tencent's short video APP WeSee during a few consecutive days. There are 10 billion samples in the dataset. In addition to labels PCR(Play Completion Ratio), PFR(PLAY Finish Rate), PSR(Play Skip Rate) as mentioned before, there are also explicit user feedback labels  LR(Like Rate), FR(Follow Rate), CMR(Comment Rate), SR(Share Rate), RCR(Read Comment Rate), RHR(Read Homepage Rate). We divide the users into three groups: new users group, low activity users group and high activity users group. Low activity users group are users with less than 60 min video watch time. 

\begin{table}[]
\renewcommand\tabcolsep{2pt}
\renewcommand\arraystretch{0.9}
  \centering
  \caption{Performance on play tasks. The improvements of MFH over \textit{baseline 3-task} are shown in the brackets.}
  \vspace{-0.5em}
    \begin{tabular}{llll}
    \multicolumn{4}{c}{\textbf{New User Group}} \\
    \toprule
    {Models} & \multicolumn{1}{l}{\begin{tabular}[c]{@{}c@{}}PCR  MSE\end{tabular}}  & \multicolumn{1}{l}{\begin{tabular}[c]{@{}c@{}}PFR   AUC \end{tabular}} & \multicolumn{1}{l}{\begin{tabular}[c]{@{}c@{}}PSR    AUC \end{tabular}} \\
    \hline 
    baseline 3-task & \makecell[l]{.5167}  & \makecell[l]{.7784}  & \makecell[l]{.7968}    \\
		flat 5-task   & .5184 & .7770 & .7973 \\
	    flat 9-task   & .5162 & .7799 & .7965 \\
		H-MTL 9-task  & .5156 & .7801 & .7989 \\
		MFH 9-task &\bf .5138$_{(\textbf{-0.56\%})}$   &\bf .7813$_{(\textbf{+0.37\%})}$   &\bf .8003$_{(\textbf{+0.44\%})}$ \\
    \bottomrule
    \vspace{-0.5em}
          &       &       &  \\
    \multicolumn{4}{c}{\textbf{Low Activity User Group}} \\
    \toprule
    {Models} & \multicolumn{1}{l}{\begin{tabular}[c]{@{}c@{}}PCR   MSE\end{tabular}}  & \multicolumn{1}{l}{\begin{tabular}[c]{@{}c@{}}PFR   AUC \end{tabular}} & \multicolumn{1}{l}{\begin{tabular}[c]{@{}c@{}}PSR    AUC \end{tabular}} \\
    \hline
    baseline 3-task  & \makecell[l]{.4807}  & \makecell[l]{.7977}  & \makecell[l]{.8179}                  \\
		flat 5-task   & .4832 & .7929 & .8174 \\
	    flat 9-task   & .4796 & .7990 & .8178 \\ 
		H-MTL 9-task  & .4792 & .8001 & .8186 \\
		MFH 9-task &\bf .4776$_{(\textbf{-0.64\%})}$  &\bf .8006$_{(\textbf{+0.36\%})}$   &\bf .8198$_{(\textbf{+0.23\%})}$  \\

    \bottomrule
    \vspace{-0.5em}
          &       &       &  \\
    \multicolumn{4}{c}{\textbf{High Activity User Group}} \\
    \toprule
    {Models} & \multicolumn{1}{l}{\begin{tabular}[c]{@{}c@{}}PCR   MSE \end{tabular}}  & \multicolumn{1}{l}{\begin{tabular}[c]{@{}c@{}}PFR   AUC \end{tabular}} & \multicolumn{1}{l}{\begin{tabular}[c]{@{}c@{}}PSR    AUC \end{tabular}} \\
    \hline
    baseline 3-task  & \makecell[l]{.4085}  & \makecell[l]{.8177}  & \makecell[l]{.8548}
		\\
		flat 5-task   & .4087 & .8170 & .8552 \\
	    flat 9-task   & .4079 & .8181 & .8536 \\
		H-MTL 9-task  & .4073 & .8183 & .8549 \\
		MFH 9-task &\bf .4070$_{(\textbf{-0.37\%})}$   &\bf .8199$_{(\textbf{+0.27\%})}$   &\bf .8563 $_{(\textbf{+0.18\%})}$  \\
    \bottomrule
    \end{tabular}%
  \label{tab:play_task}%
  \vspace{-1em}
\end{table}%

\subsubsection{Learning Tasks} There are two MTL learning models to serve the online ranking. A play tasks MTL that jointly predicts PCR, PFR and PSR, 3 tasks that are mostly related to the video playing process.  Another interactive tasks MTL that jointly predicts LR, FR, CMR, SR, RCR and RHR, the explicit user feedback behaviours.

\subsubsection{MTL Models} For the play tasks MTL, we evaluate the baseline 3-task MTL(Fig. \ref{fig_baseline_3task}), flat 5-task MTL(Fig. \ref{fig_baseline_5task}), flat 9-task MTL(Fig. \ref{fig_baseline_9task}), H-MTL 9-task(Fig. \ref{fig_hmtl}) and MFH 9-task MTL(Fig. \ref{fig_mhmtl}). For the interactive tasks MTL, we evaluate baseline 6-task, H-MTL 12-task and MFH 12-task models.

To avoid a sudden performance gap for new users or low activity users when they change group memberships, we include new user samples in all groups' task training and low activity user samples in high activity user tasks training while in serving only do inference through the corresponding user group tasks, this way the performance is smoother as user upgrade groups. As new users and low activity users are far less than high activity users, the larger user groups' training is not negatively affected. 

Since the focus in this work is the macro level task coordinate learning structures, for each MTL model, we tried different MTL switchers for the micro level shared learning such as hard sharing, MMOE, CGC\cite{tang2020progressive} and PLE and uses the performance of the best switcher choice to represent the macro level MTL structure. As a result, we adopt shared-bottom for level 0 switcher, PLE for level 1 switchers, CGC for level 2 switchers in H-MTL and MFH, and adopt PLE for switchers in rest of the models.


\begin{table*}[t]
\renewcommand\tabcolsep{2pt}
\renewcommand\arraystretch{0.9}
  \centering
  \caption{Performance on Interactive Tasks}
  \vspace{-0.5em}
    \begin{tabular}{lcccccc}
    \multicolumn{7}{c}{\textbf{New \& Low Activity User Group}} \\
    \toprule
    \multicolumn{1}{c}{Models} & LR AUC & FR AUC & CR AUC & SR AUC & RCR AUC & RHR AUC \\
    \hline
    baseline 6-task & .8821 & .9453 & .9495 & .9167 & .9113 & .8723 \\
    H-MTL 12-task & .8851(+0.34\%) & .9455(+0.02\%) & .9541(+0.48\%) & .9208(+0.45\%) & .9114(+0.01\%) & .8800(+0.88\%) \\
    MFH 12-task & .8866(\textbf{+0.51\%}) & .9461(\textbf{+0.08\%}) & .9558(\textbf{+0.66\%}) & .9239(\textbf{+0.79\%}) & .9116(\textbf{+0.03\%}) & .8882(\textbf{+1.82\%}) \\
    \bottomrule
    \vspace{-0.5em}
          &       &       &       &       &       &  \\
    \multicolumn{7}{c}{\textbf{High Activity User Group}} \\
    \toprule
    \multicolumn{1}{c}{Models} & LR AUC & FR AUC & CR AUC & SR AUC & RCR AUC & RHR AUC \\
    \hline
    baseline 6-task & .9195 & .9484 & .9722 & .9521 & .9417 & .9243 \\
    H-MTL 12-task & .9211(+0.17\%) & .9515(+0.33\%) & .9732(+0.10\%) & .9560(+0.41\%) & .9425(+0.08\%) & .9305(+0.67\%) \\
    MFH 12-task & .9221(\textbf{+0.28\%}) & .9531(\textbf{+0.50\%}) & .9758(\textbf{+0.37\%}) & .9587(\textbf{+0.70\%}) & .9434(\textbf{+0.18\%}) & .9327(\textbf{+0.91\%}) \\
    \bottomrule
    \end{tabular}%
  \label{tab:interactive_task}%
  \vspace{-1em}
\end{table*}%

\subsubsection{Experiment Setup}In the experiment, PCR prediction is a regression task trained with MSE loss and evaluated with MSE, tasks modeling other actions are all binary classification tasks trained with cross-entropy loss and evaluated with AUC. Samples in the first 14 days are used for training and the rest samples for testing. We adopt a two-layer MLP network with RELU activation and hidden layer size of [128,64] for each task's tower part in all MTL models. The experts in the switcher is implemented with a single-layer network and tune the following model-specific hyper-parameters: number of shared layers, number of experts. 

\subsubsection{Offline Evaluation with Play Task Models}Table \ref{tab:play_task} illustrates the experiment results and we mark best performance in bold. It is shown that H-MTL significantly outperform all flat task models, baseline 3-task, flat 5-task and flat 9-task in all tasks and all user divisions. With the flat task shared learning structure, introducing more tasks produces slight improvement, but much insufficient compared to the improvement H-MTL and MFH generate. H-MTL and MFH introduce new shared learning structures to better address the difference and imbalance between different user divisions. MFH further significantly outperform H-MTL in all tasks and user divisions. Of the improvement H-MTL and MFH generates, it is much more on the new users than on the active users.

We can also compare the performance between MFH 9-task and flat 9-task in Table \ref{tab:play_task}, as both models have the same 9 tasks and the only difference is MFH vs. flat on macro shared learning structure. MFH 9-task outperforms flat 9-task on all tasks: -0.46\% MSE on New\&Cmpl, -0.42\% MSE on Low\&Cmpl, -0.22\% MSE on High\&Cmpl; +0.18\% AUC on New\&Finish, +0.2\% AUC on Low\&Finish, +0.22\% AUC on High\&Finish;  +0.48\% AUC on New\&Skip, +0.24\% AUC on Low\&Skip, +0.32\% AUC on High\&Skip. This shows MFH manifest significant performance improvement over baseline flat sharing. 

Because MFH's macro structure handles large numbers of tasks much better, combined with tasks increase it generates big improvement over the flat 3-task model, better releasing the potential of large number tasks MTL.

\subsubsection{Offline Evaluation with Interactive Task Models} The interactive tasks include LR(Like Rate), FR(Follow Rate), CR (Comment Rate), SR (Share Rate), RCR(Read Comment Rate) and RHR(Read Homepage Rate) 6 tasks. For interactive task MTL, we merge the new user group and the low activity user group into one group, thus a Cartesian Product of user group and interactive behaviour types produces 12 tasks. As shown in Table \ref{tab:interactive_task}, H-MTL and MFH achieve significant improvement over the baseline model on all tasks of all user groups.

\subsubsection{Offline evaluation on 3-facets MFH}

\begin{table*}[]
\renewcommand\tabcolsep{6pt}
\renewcommand\arraystretch{1.1}
  \centering
  \caption{Performance of 3-facets MFH vs. 2-facets MFH on play tasks. The improvements of 3-facets MFH over 2-facets MFH are shown in the brackets. }
  \vspace{0.5em}
    \begin{tabular}{llll}
    \multicolumn{4}{c}{\textbf{Cold Start Items}} \\
    \toprule
    {Models} & \multicolumn{1}{l}{\begin{tabular}[c]{@{}c@{}}PCR  MSE\end{tabular}}  & \multicolumn{1}{l}{\begin{tabular}[c]{@{}c@{}}PFR   AUC \end{tabular}} & \multicolumn{1}{l}{\begin{tabular}[c]{@{}c@{}}PSR    AUC \end{tabular}} \\
    \hline  
    MFH 9-task (2facets) & \makecell[l]{.2387}  & \makecell[l]{.8471}  & \makecell[l]{.8060}    \\
	MFH 12-task (3facets)  & \bf .2363$_{(\textbf{-1.01\%})}$   & \bf .8506$_{(\textbf{+0.41\%})}$   & \bf .8091$_{(\textbf{+0.38\%})}$ \\
    \bottomrule
    \vspace{-0.5em}
          &       &       &  \\
    \multicolumn{4}{c}{\textbf{Non Cold Start Items}} \\
    \toprule
    {Models} & \multicolumn{1}{l}{\begin{tabular}[c]{@{}c@{}}PCR   MSE\end{tabular}}  & \multicolumn{1}{l}{\begin{tabular}[c]{@{}c@{}}PFR   AUC \end{tabular}} & \multicolumn{1}{l}{\begin{tabular}[c]{@{}c@{}}PSR    AUC \end{tabular}} \\
    \hline
    MFH 9-task (2facets)  & \makecell[l]{.3896}  & \makecell[l]{.8188}  & \makecell[l]{.8473}                  \\
	MFH-12task (3facets)   & \bf .3892$_{(\textbf{-0.10\%})}$  &\bf .8199$_{(\textbf{+0.13\%})}$   & \bf .8475$_{(\textbf{+0.02\%})}$  \\
    \bottomrule
    \end{tabular}%
  \label{3facetsperf}%
\end{table*}%

We conduct offline experiment to evaluate the 3-facets MFH model as shown in Fig. \ref{fig_3facets} . As new items are normally not served to new users and low activity users for a double blind cold start in practice with concerns on user experience, we only serve new cold start items to high activity users but not to low activity and new users. Thus instead of a MFH 18-task as shown in Fig 4. of the main body, we have a MFH 12-task of 3 facets, with the high activity user group's 3 tasks divided into 6 tasks, 3 for cold start content items and 3 for non cold start items. 

Same dataset as above is used to evaluate the offline performance of MFH 12-task vs. the 2 facets MFH 9 tasks. As a result, 3 facets MFH 12-task model achieves similar performance on new user and low activity users compared to 2 facets MFH 9-task model and at the same time achieves significant improvement on high activity users group where the 3rd facet is applied. As shown in Table \ref{3facetsperf}, the 3 facets MF 12-task model outperforms the 2 facets MFH 9-task model on all tasks for the high activity users, for both the non cold start items and the cold start items. A great reduction of \textbf{-1.01\%} on \textit{Cmpl} task's MSE loss are observed for cold start new items,  which is a remarkably significant improvement for content cold start as normally a 0.1\% increase of AUC or MSE already generates online metrics improvement significant enough to be observed in A/B testing.

\subsubsection{Mitigation of local overfitting}
\begin{figure}[t]
\centering
\includegraphics[width=0.6\columnwidth]{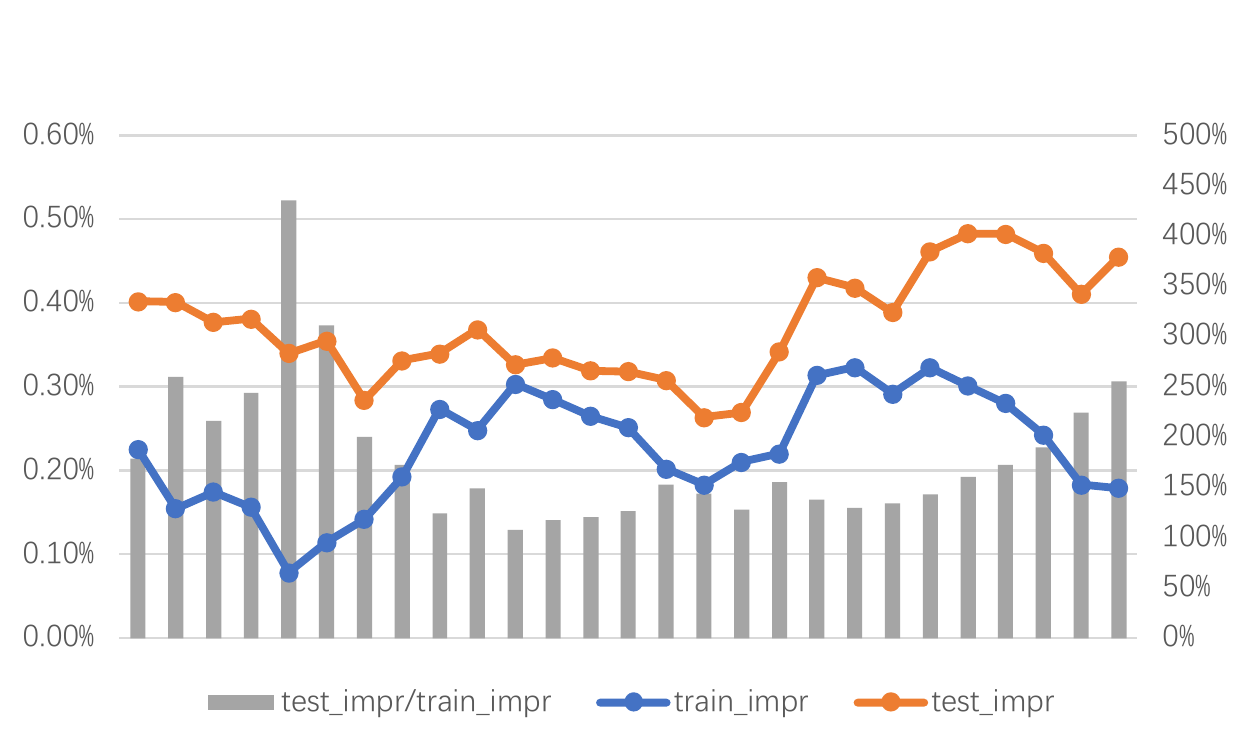}
\caption{Mitigation of local overfitting on Share Rate in Interactive Tasks MTL}
\label{fig_local_overfitting}
\end{figure}

Through multi-dimensional shared learning between the tasks, MFH maximizes the shared representation learning and mitigates the local overfitting phenomenon. Here we use the \textit{Share Rate} task in the interactive task group as an example to show the mitigation of local overfitting with MFH. As in Fig. \ref{fig_local_overfitting}, we compared MFH 12-task MTL and the baseline flat 12-task MTL on training and testing errors on new users. \textit{train\_impr} denotes the training error relative improvement from MFH 12-task to flat 12-task on Share Rate prediction, and \textit{test\_impr} denotes the testing error relative improvement accordingly. As the training time proceeds shown in Fig. \ref{fig_local_overfitting}, the test error is reduced much more than the training error, which is about 2.5 times higher. Thus, compared to the flat MTL models, MFH mitigates the local overfitting phenomenon by reducing the gap between training and testing errors.

\subsubsection{Online Evaluation} 
Online experiments are also conducted. The baseline uses flat 3-task model for the play tasks MTL and flat 6-task model for the interactive tasks MTL,  the experiment group adopts a MFH 9-task model and a MFH 12-task model accordingly. Both the experiment and the control group have the same RL fusion model adapted to the corresponding MTL ranking models. Table \ref{tab:online_results} shows the significant improvement of MFH and it is worth noting MFH shows remarkable increase of \textbf{+9.1\%} apptime per user on new users.

\begin{table}[h]
\renewcommand\tabcolsep{8pt}
\renewcommand\arraystretch{0.9}
  \centering
  \caption{Online Experiment Evaluation}
  \vspace{-0.5em}
    \begin{tabular}{lll}
    \multicolumn{3}{c}{\textbf{Online Evaluation}} \\
    \toprule
    {Models} & \multicolumn{1}{l}{\begin{tabular}[c]{@{}c@{}}apptime\end{tabular}}  & \multicolumn{1}{l}{\begin{tabular}[c]{@{}c@{}}2nd-day retention \end{tabular}} \\
    \hline 
    Baseline Flat MTLs & \makecell[l]{-}  & \makecell[l]{-}      \\
		MFH   & +2.14\%   & +0.19\%    \\
		MFH on new users   & +9.1\%   & +1.85\%    \\
    \bottomrule
    \end{tabular}%
  \label{tab:online_results}%
  \vspace{-1em}
\end{table}%

\begin{table*}[t]
\renewcommand\tabcolsep{2pt}
\renewcommand\arraystretch{1.0}
  \centering
  \caption{Performance on Ali-CCP Dateset}
  \vspace{0.5em}
    \begin{tabular}{lllllll}
    \toprule
    \multirow{2}[2]{*}{} & \multicolumn{2}{l}{Low Purchase User Group} & \multicolumn{2}{l}{Medium Purchase User Group} & \multicolumn{2}{l}{High Purchase User Group} \\
          & CTR AUC & CVR AUC & CTR AUC & CVR AUC & CTR AUC & CVR AUC \\
    \hline
    Baseline 2-task & .6139  & .6457  & .6151  & .6408  & .6120  & .6439  \\
    Baseline 6-task & .6140$_{(+0.02\%)}$ & .6466$_{(+0.14\%)}$ & .6160$_{(+0.15\%)}$ & .6541$_{(\textbf{+2.08\%})}$ & .6135$_{(\textbf{+0.25\%})}$ & .6446$_{(+0.11\%)}$ \\
    MFH 6-task & .6154$_{(\textbf{+0.24\%})}$ & .6494$_{(\textbf{+0.57\%})}$ & .6179$_{(\textbf{+0.46\%})}$ & .6544$_{(\textbf{+2.12\%})}$ & .6144$_{(\textbf{+0.39\%})}$ & .6457$_{(\textbf{+0.28\%})}$ \\
    \bottomrule
    \end{tabular}%
  \label{tab:ali-cpp}%
\end{table*}%

\subsection{\textbf{Evaluation on Public Dataset}}
In this subsection, experiments are conducted on public benchmark datasets to evaluate MFH in more application scenarios.

\subsubsection{Dataset} 
Ali-CCP(Alibaba Click and Conversion Prediction) Dataset is a public dataset extracted from Taobao’s Recommender System. The dataset includes 84 million data samples equally divided into training set and testing set, which contain 3.4 million clicks and 18 thousand conversions. CTR(Click Through Rate) and CVR (Conversion Rate) are two tasks modeling user actions of click and purchase in the dataset.

\subsubsection{Experiment Setup}
For users in the dataset, we define \textit{Purchasing Power} = CVR / CTR to characterize the user tendencies of purchasement. Combined with the User Consumption Level Type I and II fields provided in the dataset, users are clustered into three groups, low, medium and high purchasing power user groups, containing 40\%, 35\% and 25\% users respectively. Since there are a CTR task and a CVR task for every group, we have a total of 6 tasks. In our experiment, a flat 2-task and flat 6-task Model(PLE as switcher) are used as the baseline models, compared with a MFH 6-task Model. For each task in both models, we adopt a two-layer MLP network with hidden layer size of [64, 32].

\begin{figure}[]
\centering
\includegraphics[width=0.99\columnwidth]{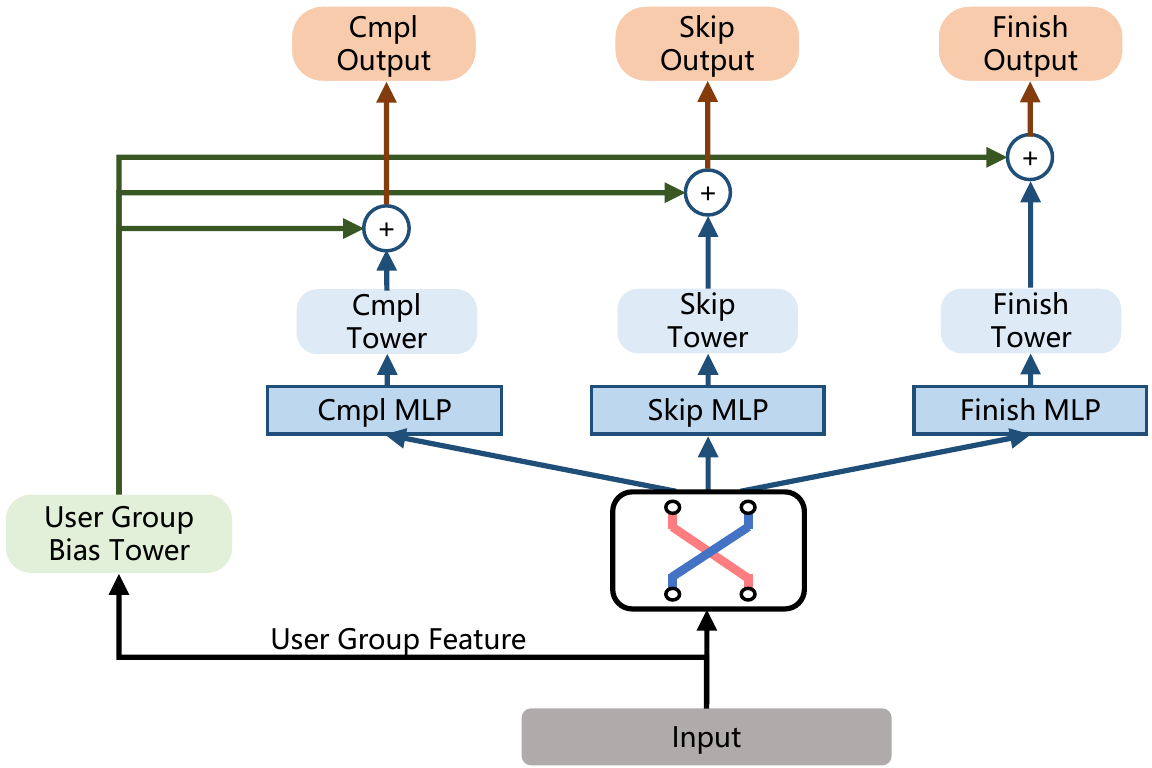}
\caption{biasnet}
\label{fig_biasnet}
\end{figure}

\subsubsection{Experiment Results} 
As shown in Table \ref{tab:ali-cpp}, compared to the baseline 2-task and baseline 6-task Models, MFH significantly improves the AUC of all six tasks. And the baseline 6-task model performs slightly better than the baseline 2-task model.

\subsection{Comparison With Biasnet}
Biasnet\cite{moore2018modeling,zhao2019recommending} is a natural alternative to address the user group differences with a side shallow bias tower that outputs a bias logit to be combined with the main tower. Viewing the neural network as a network learning a nonlinear mapping from the input feature space to the labels, the MTL approach provides a more general mapping than the biasnet approach as the task specific towers provide greater flexibility in fitting the label. Biasnet uses a stronger induction bias which may be beneficial for cases where the bias difference can be modeled with a few layers of simple neural mappings and the biased feature value is continuous. Thus, the MTL approach is a more general framework, it can actually include the biasnet structure as part of the MTL network. Offline evaluation is also conducted to compare MFH's performance with biasnet. As shown Fig. \ref{fig_biasnet}, we replace the user group facet with a bias tower that takes the user group feature as its input, and the bias output is added to the logits of task towers. The bias tower is composed of two-layer MLP with RELU activation and hidden layer size of [128,64].

\begin{table}[]
\renewcommand\tabcolsep{6pt}
\renewcommand\arraystretch{1.1}
  \centering
  \caption{Performance of MFH vs. Biasnet on play tasks. The improvements of MFH over \textit{Biasnet} are shown in the brackets.}
  \vspace{0.5em}
    \begin{tabular}{llll}
    \multicolumn{4}{c}{\textbf{New User Group}} \\
    \toprule
    {Models} & \multicolumn{1}{l}{\begin{tabular}[c]{@{}c@{}}PCR  MSE\end{tabular}}  & \multicolumn{1}{l}{\begin{tabular}[c]{@{}c@{}}PFR   AUC \end{tabular}} & \multicolumn{1}{l}{\begin{tabular}[c]{@{}c@{}}PSR    AUC \end{tabular}} \\
    \hline  
    Biasnet & \makecell[l]{.4457}  & \makecell[l]{.7767}  & \makecell[l]{.7914}    \\
		MFH 9-task  & \bf .4423$_{(\textbf{-0.77\%})}$   & \bf .7773$_{(\textbf{+0.08\%})}$   & \bf .7926$_{(\textbf{+0.15\%})}$ \\
    \bottomrule
    \multicolumn{4}{c}{\textbf{Low Activity User Group}} \\
    \toprule
    {Models} & \multicolumn{1}{l}{\begin{tabular}[c]{@{}c@{}}PCR   MSE\end{tabular}}  & \multicolumn{1}{l}{\begin{tabular}[c]{@{}c@{}}PFR   AUC \end{tabular}} & \multicolumn{1}{l}{\begin{tabular}[c]{@{}c@{}}PSR    AUC \end{tabular}} \\
    \hline
    Biasnet  & \makecell[l]{.4327}  & \makecell[l]{\bf .8019}  & \makecell[l]{.8041}                  \\
		MFH 9-task   & \bf .4276$_{(\textbf{-1.19\%})}$  &  .8011$_{(-0.10\%)}$   & \bf .8051$_{(\textbf{+0.12\%})}$  \\
    \bottomrule
    \multicolumn{4}{c}{\textbf{High Activity User Group}} \\
    \toprule
    {Models} & \multicolumn{1}{l}{\begin{tabular}[c]{@{}c@{}}PCR   MSE \end{tabular}}  & \multicolumn{1}{l}{\begin{tabular}[c]{@{}c@{}}PFR   AUC \end{tabular}} & \multicolumn{1}{l}{\begin{tabular}[c]{@{}c@{}}PSR    AUC \end{tabular}} \\
    \hline
    Biasnet  & \makecell[l]{.3829}  & \makecell[l]{\bf .8201}  & \makecell[l]{.8497} 
		\\
		MFH 9-task  &  \bf .3799$_{(\textbf{-0.79\%})}$ &  .8191$_{(-0.12\%)}$   & \bf .8504$_{(\textbf{+0.08\%})}$  \\
    \bottomrule
    \end{tabular}%
  \label{mfh-vs-biasnet}%
\end{table}%

As Table \ref{mfh-vs-biasnet} shows, the 2 facets MFH 9-task MTL outperforms Biasnet on the \textit{Cmpl} task's PCR(Predicted Completion Ratio) MSE loss for all user groups, and also outperforms Biasnet in all tasks for new users, slightly underperforms Biasnet on PFR(Predicted Finish Rate) and PSR(Predicted Skip Rate) on partial user groups. As the PCR task and new user group are more important, overall MFH performs better than biasnet.

\section{Conclusion}
In this paper we propose a novel MTL model called Multi-Faceted Hierarchical multi-task learning model(MFH), which uses a multi-faceted hierarchical tree structure to improve the MTL efficiency and scalability from the macro perspective of task sharing. Offline and online experiment results on the industrial and public datasets show significant and consistent improvements of MFH over baseline SOTA models. Researching the possibilities of applying Meta Learning on the MFH tree structures will be the focus of future work.


\bibliographystyle{IEEEtran}
\bibliography{IEEEabrv,references}

\begin{thebibliography}{10}
\providecommand{\url}[1]{#1}
\csname url@samestyle\endcsname
\providecommand{\newblock}{\relax}
\providecommand{\bibinfo}[2]{#2}
\providecommand{\BIBentrySTDinterwordspacing}{\spaceskip=0pt\relax}
\providecommand{\BIBentryALTinterwordstretchfactor}{4}
\providecommand{\BIBentryALTinterwordspacing}{\spaceskip=\fontdimen2\font plus
\BIBentryALTinterwordstretchfactor\fontdimen3\font minus
  \fontdimen4\font\relax}
\providecommand{\BIBforeignlanguage}[2]{{%
\expandafter\ifx\csname l@#1\endcsname\relax
\typeout{** WARNING: IEEEtran.bst: No hyphenation pattern has been}%
\typeout{** loaded for the language `#1'. Using the pattern for}%
\typeout{** the default language instead.}%
\else
\language=\csname l@#1\endcsname
\fi
#2}}
\providecommand{\BIBdecl}{\relax}
\BIBdecl

\bibitem{tang2020progressive}
H.~Tang, J.~Liu, M.~Zhao, and X.~Gong, ``Progressive layered extraction (ple):
  A novel multi-task learning (mtl) model for personalized recommendations,''
  in \emph{Fourteenth ACM Conference on Recommender Systems}, 2020, pp.
  269--278.

\bibitem{ma2018modeling}
J.~Ma, Z.~Zhao, X.~Yi, J.~Chen, L.~Hong, and E.~H. Chi, ``Modeling task
  relationships in multi-task learning with multi-gate mixture-of-experts,'' in
  \emph{Proceedings of the 24th ACM SIGKDD International Conference on
  Knowledge Discovery \& Data Mining}, 2018, pp. 1930--1939.

\bibitem{caruana1997multitask}
R.~Caruana, ``Multitask learning,'' \emph{Machine learning}, vol.~28, no.~1,
  pp. 41--75, 1997.

\bibitem{ZhangYangMTLSurvey}
Y.~Zhang and Q.~Yang, ``A survey on multi-task learning,'' \emph{IEEE
  Transactions on Knowledge and Data Engineering}, pp. 1--1, 2021.

\bibitem{Davidson10theyoutube}
J.~Davidson, P.~Nandy, B.~Liebald, T.~V. Vleet, and J.~Liu, ``The youtube video
  recommendation system,'' in \emph{In Proceedings of the fourth ACM conference
  on Recommender systems, RecSys ’10}.\hskip 1em plus 0.5em minus 0.4em\relax
  ACM, 2010, pp. 293--296.

\bibitem{bansal2016ask}
T.~Bansal, D.~Belanger, and A.~McCallum, ``Ask the gru: Multi-task learning for
  deep text recommendations,'' in \emph{proceedings of the 10th ACM Conference
  on Recommender Systems}, 2016, pp. 107--114.

\bibitem{felfernig2018group}
A.~Felfernig, L.~Boratto, M.~Stettinger, and M.~Tkal{\v{c}}i{\v{c}},
  \emph{Group recommender systems: An introduction}.\hskip 1em plus 0.5em minus
  0.4em\relax Springer, 2018.

\bibitem{collobert2008unified}
R.~Collobert and J.~Weston, ``A unified architecture for natural language
  processing: Deep neural networks with multitask learning,'' in
  \emph{Proceedings of the 25th international conference on Machine learning},
  2008, pp. 160--167.

\bibitem{sanh2019hierarchical}
V.~Sanh, T.~Wolf, and S.~Ruder, ``A hierarchical multi-task approach for
  learning embeddings from semantic tasks,'' in \emph{Proceedings of the AAAI
  Conference on Artificial Intelligence}, vol.~33, no.~01, 2019, pp.
  6949--6956.

\bibitem{nguyen2019multi}
D.-K. Nguyen and T.~Okatani, ``Multi-task learning of hierarchical
  vision-language representation,'' in \emph{Proceedings of the IEEE Conference
  on Computer Vision and Pattern Recognition}, 2019, pp. 10\,492--10\,501.

\bibitem{fan2017hd}
J.~Fan, T.~Zhao, Z.~Kuang, Y.~Zheng, J.~Zhang, J.~Yu, and J.~Peng, ``Hd-mtl:
  Hierarchical deep multi-task learning for large-scale visual recognition,''
  \emph{IEEE transactions on image processing}, vol.~26, no.~4, pp. 1923--1938,
  2017.

\bibitem{kendall2018multi}
A.~Kendall, Y.~Gal, and R.~Cipolla, ``Multi-task learning using uncertainty to
  weigh losses for scene geometry and semantics,'' in \emph{Proceedings of the
  IEEE conference on computer vision and pattern recognition}, 2018, pp.
  7482--7491.

\bibitem{misra2016cross}
I.~Misra, A.~Shrivastava, A.~Gupta, and M.~Hebert, ``Cross-stitch networks for
  multi-task learning,'' in \emph{Proceedings of the IEEE conference on
  computer vision and pattern recognition}, 2016, pp. 3994--4003.

\bibitem{ruder2017sluice}
S.~Ruder, J.~Bingel, I.~Augenstein, and A.~S{\o}gaard, ``Sluice networks:
  Learning what to share between loosely related tasks,'' \emph{arXiv preprint
  arXiv:1705.08142}, vol.~2, 2017.

\bibitem{jacobs1991adaptive}
R.~A. Jacobs, M.~I. Jordan, S.~J. Nowlan, and G.~E. Hinton, ``Adaptive mixtures
  of local experts,'' \emph{Neural computation}, vol.~3, no.~1, pp. 79--87,
  1991.

\bibitem{xie2021real}
R.~Xie, R.~Wang, S.~Zhang, Z.~Yang, F.~Xia, and L.~Lin, ``Real-time relevant
  recommendation suggestion,'' in \emph{Proceedings of the 14th ACM
  International Conference on Web Search and Data Mining}, 2021, pp. 112--120.

\bibitem{ding2021mssm}
K.~Ding, X.~Dong, Y.~He, L.~Cheng, C.~Fu, Z.~Huan, H.~Li, T.~Yan, L.~Zhang,
  X.~Zhang \emph{et~al.}, ``Mssm: A multiple-level sparse sharing model for
  efficient multi-task learning,'' in \emph{Proceedings of the 44th
  International ACM SIGIR Conference on Research and Development in Information
  Retrieval}, 2021, pp. 2237--2241.

\bibitem{ma2018entire}
X.~Ma, L.~Zhao, G.~Huang, Z.~Wang, Z.~Hu, X.~Zhu, and K.~Gai, ``Entire space
  multi-task model: An effective approach for estimating post-click conversion
  rate,'' in \emph{The 41st International ACM SIGIR Conference on Research \&
  Development in Information Retrieval}, 2018, pp. 1137--1140.

\bibitem{wen2021hierarchically}
H.~Wen, J.~Zhang, F.~Lv, W.~Bao, T.~Wang, and Z.~Chen, ``Hierarchically
  modeling micro and macro behaviors via multi-task learning for conversion
  rate prediction,'' \emph{arXiv preprint arXiv:2104.09713}, 2021.

\bibitem{zoph2016neural}
B.~Zoph and Q.~V. Le, ``Neural architecture search with reinforcement
  learning,'' \emph{arXiv preprint arXiv:1611.01578}, 2016.

\bibitem{rosenbaum2017routing}
C.~Rosenbaum, T.~Klinger, and M.~Riemer, ``Routing networks: Adaptive selection
  of non-linear functions for multi-task learning,'' \emph{arXiv preprint
  arXiv:1711.01239}, 2017.

\bibitem{ma2019snr}
J.~Ma, Z.~Zhao, J.~Chen, A.~Li, L.~Hong, and E.~H. Chi, ``Snr: Sub-network
  routing for flexible parameter sharing in multi-task learning,'' in
  \emph{Proceedings of the AAAI Conference on Artificial Intelligence},
  vol.~33, no.~01, 2019, pp. 216--223.

\bibitem{chen2020boosting}
X.~Chen, X.~Gu, and L.~Fu, ``Boosting share routing for multi-task learning,''
  \emph{arXiv preprint arXiv:2009.00387}, 2020.

\bibitem{lawrence2000overfitting}
S.~Lawrence and C.~L. Giles, ``Overfitting and neural networks: conjugate
  gradient and backpropagation,'' in \emph{Proceedings of the IEEE-INNS-ENNS
  International Joint Conference on Neural Networks. IJCNN 2000. Neural
  Computing: New Challenges and Perspectives for the New Millennium},
  vol.~1.\hskip 1em plus 0.5em minus 0.4em\relax IEEE, 2000, pp. 114--119.

\bibitem{jiang2006svm}
F.~Jiang, \emph{SVM-based negative data mining to binary classification}.\hskip
  1em plus 0.5em minus 0.4em\relax Georgia State University, 2006.

\bibitem{webster2019detecting}
R.~Webster, J.~Rabin, L.~Simon, and F.~Jurie, ``Detecting overfitting of deep
  generative networks via latent recovery,'' in \emph{Proceedings of the
  IEEE/CVF Conference on Computer Vision and Pattern Recognition}, 2019, pp.
  11\,273--11\,282.

\bibitem{covington2016deep}
P.~Covington, J.~Adams, and E.~Sargin, ``Deep neural networks for youtube
  recommendations,'' in \emph{Proceedings of the 10th ACM conference on
  recommender systems}, 2016, pp. 191--198.

\bibitem{moore2018modeling}
J.~Moore, J.~Pfeiffer, K.~Wei, R.~Iyer, D.~Charles, R.~Gilad-Bachrach,
  L.~Boyles, and E.~Manavoglu, ``Modeling and simultaneously removing bias via
  adversarial neural networks,'' \emph{arXiv preprint arXiv:1804.06909}, 2018.

\bibitem{zhao2019recommending}
Z.~Zhao, L.~Hong, L.~Wei, J.~Chen, A.~Nath, S.~Andrews, A.~Kumthekar,
  M.~Sathiamoorthy, X.~Yi, and E.~Chi, ``Recommending what video to watch next:
  a multitask ranking system,'' in \emph{Proceedings of the 13th ACM Conference
  on Recommender Systems}, 2019, pp. 43--51.

\end{thebibliography}

\end{document}